\documentclass{article}

\usepackage{arxiv}

\usepackage[utf8]{inputenc} 
\usepackage[T1]{fontenc}    
\usepackage{hyperref}       
\usepackage{url}            
\usepackage{booktabs}       
\usepackage{amsfonts}       
\usepackage{nicefrac}       
\usepackage{microtype}      
\usepackage{lipsum}

\usepackage{color}
\usepackage{booktabs,threeparttable}
\usepackage{colortbl}
\usepackage[export]{adjustbox}
\usepackage{caption}

\title{CSPNet: A New Backbone that can Enhance Learning Capability of CNN}

\author{
  Chien-Yao Wang \\
  Institute of Information Science\\
  Academia Sinica, Taiwan\\
  \texttt{x102432003@yahoo.com.tw} \\
  \And
  Hong-Yuan Mark Liao \\
  Institute of Information Science\\
  Academia Sinica, Taiwan\\
  \texttt{liao@iis.sinica.edu.tw} \\
  \And
  I-Hau Yeh \\
  Elan Microelectronics Corporation, Taiwan\\
  \texttt{ihyeh@emc.com.tw} \\
  \And
  Yueh-Hua Wu\thanks{Corresponding author.} \\
  Institute of Information Science\\
  Academia Sinica, Taiwan\\
  \texttt{kriswu@iis.sinica.edu.tw} \\
  \And
  Ping-Yang Chen \\
  Department of Computer Science\\
  National Chiao Tung University, Taiwan\\
  \texttt{pingyang.cs08g@nctu.edu.tw} \\
  \And
  Jun-Wei Hsieh \\
  College of Artificial Intelligence and Green Energy\\
  National Chiao Tung University, Taiwan\\
  \texttt{jwhsieh@nctu.edu.tw} \\
}

\begin{document}
\maketitle

\begin{abstract}
   Neural networks have enabled state-of-the-art approaches to achieve incredible results on computer vision tasks such as object detection.  However, such success greatly relies on costly computation resources, which hinders people with cheap devices from appreciating the advanced technology.  In this paper, we propose Cross Stage Partial Network (CSPNet) to mitigate the problem that previous works require heavy inference computations from the network architecture perspective.  We attribute the problem to the duplicate gradient information within network optimization.  The proposed networks respect the variability of the gradients by integrating feature maps from the beginning and the end of a network stage, which, in our experiments, reduces computations by 20\% with equivalent or even superior accuracy on the ImageNet dataset, and significantly outperforms state-of-the-art approaches in terms of AP$_{50}$ on the MS COCO object detection dataset.  The CSPNet is easy to implement and general enough to cope with architectures based on ResNet, ResNeXt, and DenseNet. Source code is at \url{https://github.com/WongKinYiu/CrossStagePartialNetworks}.
\end{abstract}


\section{Introduction}

Neural networks have been shown to be especially powerful when it gets deeper \cite{he2016deep, xie2017aggregated, huang2017densely} and wider \cite{Zagoruyko2016WRN}. However, extending the architecture of neural networks usually brings up a lot more computations, which makes computationally heavy tasks such as object detection unaffordable for most people. Light-weight computing has gradually received stronger attention since real-world applications usually require short inference time on small devices, which poses a serious challenge for computer vision algorithms. Although some approaches were designed exclusively for mobile CPU \cite{howard2017mobilenets, sandler2018mobilenetv2, howard2019searching, tan2019mnasnet, zhang2018shufflenet, ma2018shufflenetv2}, the depth-wise separable convolution techniques they adopted are not compatible with industrial IC design such as Application-Specific Integrated Circuit (ASIC) for edge-computing systems.
In this work, we investigate the computational burden in state-of-the-art approaches such as ResNet, ResNeXt, and DenseNet. We further develop computationally efficient components that enable the mentioned networks to be deployed on both CPUs and mobile GPUs without sacrificing the performance.

\begin{figure}[t!]
	\begin{center}
		\includegraphics[width=0.6\linewidth]{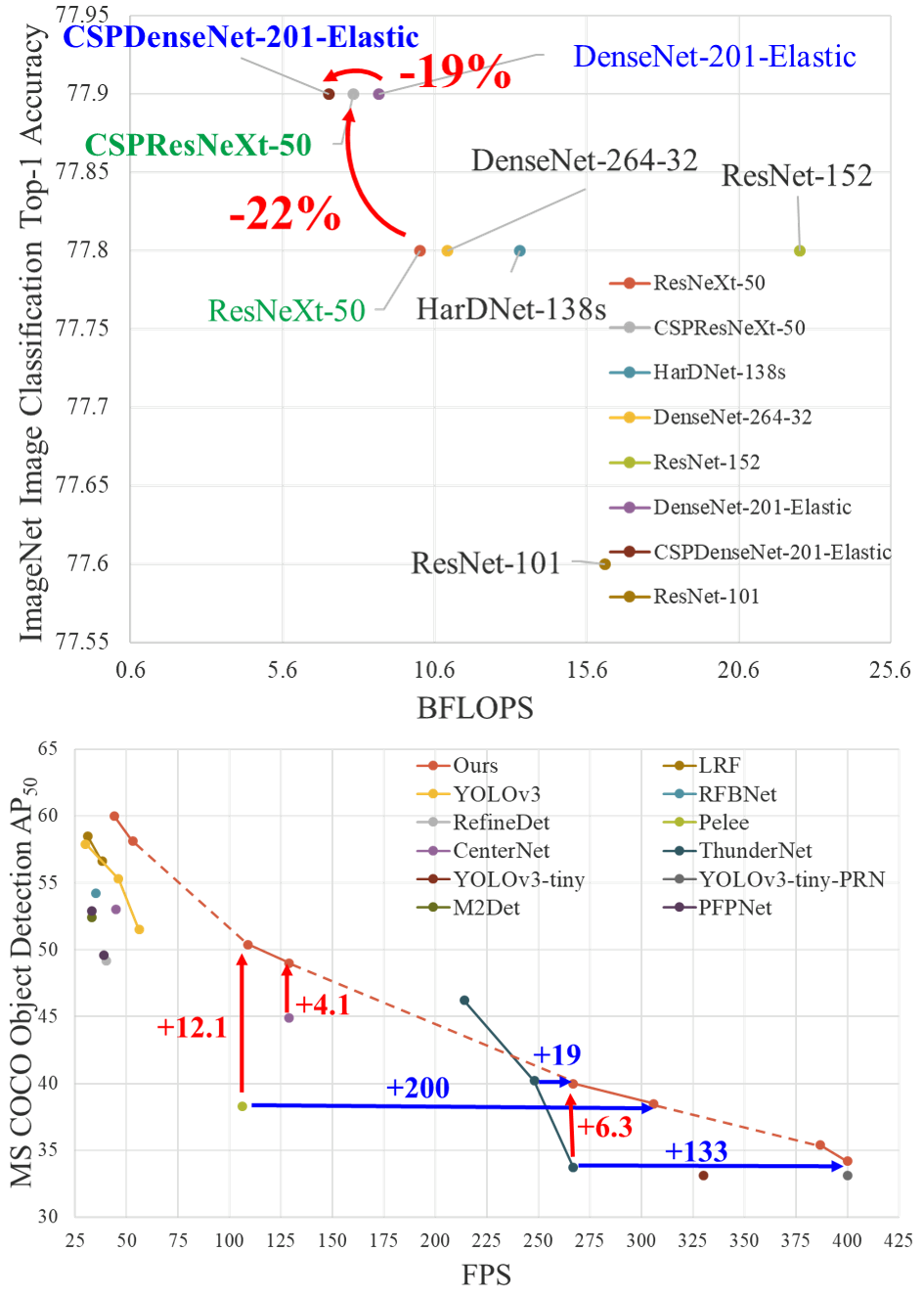}
	\end{center}
	\caption{Proposed CSPNet can be applied on ResNet \cite{he2016deep}, ResNeXt \cite{xie2017aggregated}, DenseNet \cite{huang2017densely}, etc.  It not only reduce computation cost and memory usage of these networks, but also benefit on inference speed and accuracy.  }
	\label{fig:compare}
\end{figure}

\begin{figure*}[t]
	\begin{center}
		\includegraphics[width=0.98\linewidth]{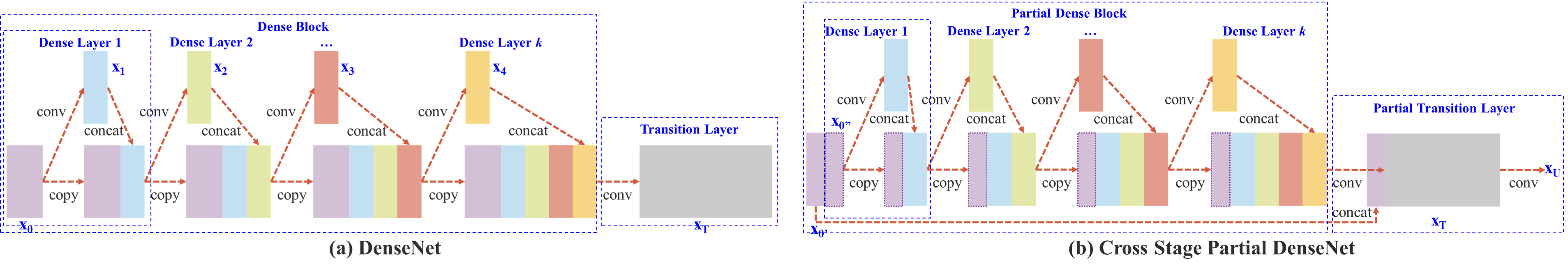}
	\end{center}
	\caption{Illustrations of (a) DenseNet and (b) our proposed Cross Stage Partial DenseNet (CSPDenseNet).  CSPNet separates feature map of the base layer into two part, one part will go through a dense block and a transition layer; the other one part is then combined with transmitted feature map to the next stage.  }
	\label{fig:cspnet}
\end{figure*}

In this study, we introduce Cross Stage Partial Network (CSPNet).  The main purpose of designing CSPNet is to enable this architecture to achieve a richer gradient combination while reducing the amount of computation.  This aim is achieved by partitioning feature map of the base layer into two parts and then merging them through a proposed cross-stage hierarchy.  Our main concept is to make the gradient flow propagate through different network paths by splitting the gradient flow.  In this way, we have confirmed that the propagated gradient information can have a large correlation difference by switching concatenation and transition steps.  In addition, CSPNet can greatly reduce the amount of computation, and improve inference speed as well as accuracy, as illustrated in Fig \ref{fig:compare}.  The proposed CSPNet-based object detector deals with the following three problems:

\textit{\textbf{1) Strengthening learning ability of a CNN }} The accuracy of existing CNN is greatly degraded after lightweightening, so we hope to strengthen CNN's learning ability, so that it can maintain sufficient accuracy while being lightweightening.  The proposed CSPNet can be easily applied to ResNet, ResNeXt, and DenseNet.  After applying CSPNet on the above mentioned networks, the computation effort can be reduced from 10\% to 20\%, but it outperforms ResNet \cite{he2016deep}, ResNeXt \cite{xie2017aggregated}, DenseNet \cite{huang2017densely}, HarDNet \cite{chao2019hardnet}, Elastic \cite{wang2019elastic}, and Res2Net \cite{gao2019res2net}, in terms of accuracy, in conducting image classification task on ImageNet \cite{deng2009imagenet}.

\textit{\textbf{2) Removing computational bottlenecks }} Too high a computational bottleneck will result in more cycles to complete the inference process, or some arithmetic units will often idle.  Therefore, we hope we can evenly distribute the amount of computation at each layer in CNN so that we can effectively upgrade the utilization rate of each computation unit and thus reduce unnecessary energy consumption.  It is noted that the proposed CSPNet makes the computational bottlenecks of PeleeNet \cite{wang2018pelee} cut into half.  Moreover, in the MS COCO \cite{lin2014microsoft} dataset-based object detection experiments, our proposed model can effectively reduce 80\% computational bottleneck when test on YOLOv3-based models.

\textit{\textbf{3) Reducing memory costs }} The wafer fabrication cost of Dynamic Random-Access Memory (DRAM) is very expensive, and it also takes up a lot of space.  If one can effectively reduce the memory cost, he/she will greatly reduce the cost of ASIC.  In addition, a small area wafer can be used in a variety of edge computing devices.  In reducing the use of memory usage, we adopt cross-channel pooling \cite{goodfellow2013maxout} to compress the feature maps during the feature pyramid generating process.  In this way, the proposed CSPNet with the proposed object detector can cut down 75\% memory usage on PeleeNet when generating feature pyramids.

Since CSPNet is able to promote the learning capability of a CNN, we thus use smaller models to achieve better accuracy.  Our proposed model can achieve 50\% COCO AP$_{50}$ at 109 fps on GTX 1080ti.  Since CSPNet can effectively cut down a significant amount of memory traffic, our proposed method can achieve 40\% COCO AP$_{50}$ at 52 fps on Intel Core i9-9900K.  In addition, since CSPNet can significantly lower down the computational bottleneck and Exact Fusion Model (EFM) can effectively cut down the required memory bandwidth, our proposed method can achieve 42\% COCO AP$_{50}$ at 49 fps on Nvidia Jetson TX2.

\section{Related work}

{\bf CNN architectures design.} In ResNeXt \cite{xie2017aggregated}, Xie \textit{et al.} first demonstrate that cardinality can be more effective than the dimensions of width and depth.  DenseNet \cite{huang2017densely} can significantly reduce the number of parameters and computations due to the strategy of adopting a large number of reuse features.  And it concatenates the output features of all preceding layers as the next input, which can be considered as the way to maximize cardinality.  SparseNet \cite{zhu2018sparsely} adjusts dense connection to exponentially spaced connection can effectively improve parameter utilization and thus result in better outcomes.  Wang \textit{et al.} further explain why high cardinality and sparse connection can improve the learning ability of the network by the concept of gradient combination and developed the partial ResNet (PRN) \cite{wang2019enriching}.  For improving the inference speed of CNN, Ma \textit{et al.} \cite{ma2018shufflenetv2} introduce four guidelines to be followed and design ShuffleNet-v2.  Chao \textit{et al.} \cite{chao2019hardnet} proposed a low memory traffic CNN called Harmonic DenseNet (HarDNet) and a metric Convolutional Input/Output (CIO) which is an approximation of DRAM traffic proportional to the real DRAM traffic measurement.

{\bf Real-time object detector.} The most famous two real-time object detectors are YOLOv3 \cite{redmon2018yolov3} and SSD \cite{liu2016ssd}.  Based on SSD, LRF \cite{wang2019learning} and RFBNet \cite{liu2018receptive} can achieve state-of-the-art real-time object detection performance on GPU.  Recently, anchor-free based object detector \cite{duan2019centernet, zhou2019objects, law2018cornernet, law2019cornernet, zhang2019freeanchor} has become main-stream object detection system.  Two object detector of this sort are CenterNet \cite{zhou2019objects} and CornerNet-Lite \cite{law2019cornernet}, and they both perform very well in terms of efficiency and efficacy.  For real-time object detection on CPU or mobile GPU, SSD-based Pelee \cite{wang2018pelee}, YOLOv3-based PRN \cite{wang2019enriching}, and Light-Head RCNN \cite{li2017light}-based ThunderNet \cite{qin2019thundernet} all receive excellent performance on object detection.

\section{Method}

\subsection{Cross Stage Partial Network}
\label{subsec:cspnet}

{\bf DenseNet.} Figure \ref{fig:cspnet} (a) shows the detailed structure of one-stage of the DenseNet proposed by Huang \textit{et al.} \cite{huang2017densely}.  Each stage of a DenseNet contains a dense block and a transition layer, and each dense block is composed of $k$ dense layers.  The output of the $i^{th}$ dense layer will be concatenated with the input of the $i^{th}$ dense layer, and the concatenated outcome will become the input of the $(i+1)^{th}$ dense layer.  The equations showing the above-mentioned mechanism can be expressed as:

\begin{equation}
\label{equation:densenet}
\includegraphics[valign=c,scale=0.5]{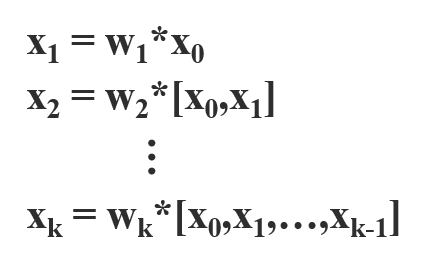}
\end{equation}
where $*$ represents the convolution operator, and $[x_{0},x_{1},...]$ means to concatenate $x_{0},x_{1},...$, and $w_{i}$ and $x_{i}$ are the weights and output of the  $i^{th}$ dense layer, respectively.

If one makes use of a backpropagation algorithm to update weights, the equations of weight updating can be written as:

\begin{equation}
\label{equation:densenetg}
\includegraphics[valign=c,scale=0.5]{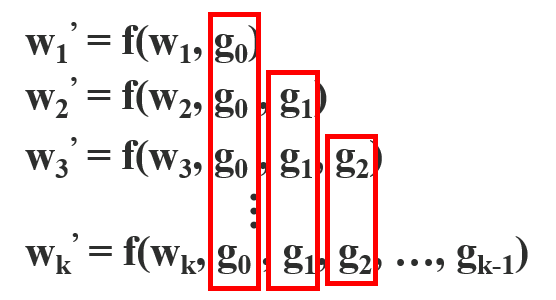}
\end{equation}
where $f$ is the function of weight updating, and $g_{i}$ represents the gradient propagated to the $i^{th}$ dense layer.  We can find that large amount of gradient information are reused for updating weights of different dense layers.  This will result in different dense layers repeatedly learn copied gradient information.

{\bf Cross Stage Partial DenseNet.} The architecture of one-stage of the proposed CSPDenseNet is shown in Figure \ref{fig:cspnet} (b).  A stage of CSPDenseNet is composed of a partial dense block and a partial transition layer.  In a partial dense block, the feature maps of the base layer in a stage are split into two parts through channel $x_{0}=[x_{0}^{'}, x_{0}^{''}]$.  Between $x_{0}^{''}$ and $x_{0}^{'}$, the former is directly linked to the end of the stage, and the latter will go through a dense block.  All steps involved in a partial transition layer are as follows: First, the output of dense layers, $[x_{0}^{''}, x_{1}, ..., x_{k}]$, will undergo a transition layer.  Second, the output of this transition layer, $x_{T}$, will be concatenated with $x_{0}^{''}$ and undergo another transition layer, and then generate output $x_{U}$.  The equations of feed-forward pass and weight updating of CSPDenseNet are shown in Equations \ref{equation:cspdensenet} and \ref{equation:cspdensenetg}, respectively.

\begin{equation}
\label{equation:cspdensenet}
\includegraphics[valign=c,scale=0.5]{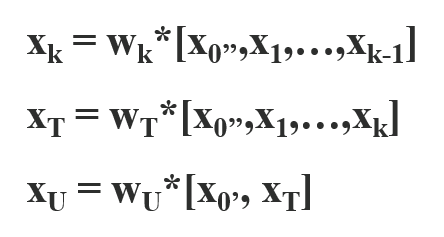}
\end{equation}

\begin{equation}
\label{equation:cspdensenetg}
\includegraphics[valign=c,scale=0.5]{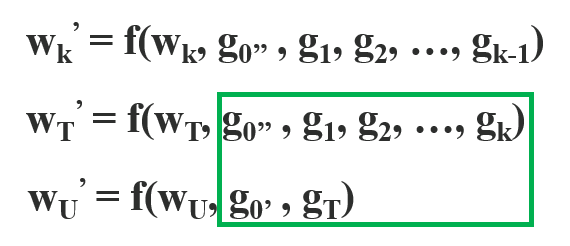}
\end{equation}

We can see that the gradients coming from the dense layers are separately integrated.  On the other hand, the feature map $x_{0}^{'}$ that did not go through the dense layers is also separately integrated.  As to the gradient information for updating weights, both sides do not contain duplicate gradient information that belongs to other sides.  

Overall speaking, the proposed CSPDenseNet preserves the advantages of DenseNet's feature reuse characteristics, but at the same time prevents an excessively amount of duplicate gradient information by truncating the gradient flow.  This idea is realized by designing a hierarchical feature fusion strategy and used in a partial transition layer.

{\bf Partial Dense Block.} The purpose of designing partial dense blocks is to \textit{1.) increase gradient path}: Through the split and merge strategy, the number of gradient paths can be doubled.  Because of the cross-stage strategy, one can alleviate the disadvantages caused by using explicit feature map copy for concatenation; \textit{2.) balance computation of each layer}: usually, the channel number in the base layer of a DenseNet is much larger than the growth rate.  Since the base layer channels involved in the dense layer operation in a partial dense block account for only half of the original number, it can effectively solve nearly half of the computational bottleneck; and \textit{3.) reduce memory traffic}: Assume the base feature map size of a dense block in a DenseNet is $w \times h \times c$, the growth rate is $d$, and there are in total $m$ dense layers.  Then, the CIO of that dense block is $(c \times m) + ((m^{2}+m) \times d)/2$, and the CIO of partial dense block is $((c \times m) + (m^{2}+m) \times d)/2$.  While $m$ and $d$ are usually far smaller than $c$, a partial dense block is able to save at most half of the memory traffic of a network.

\begin{figure}[h]
	\begin{center}
		\includegraphics[width=0.4\linewidth]{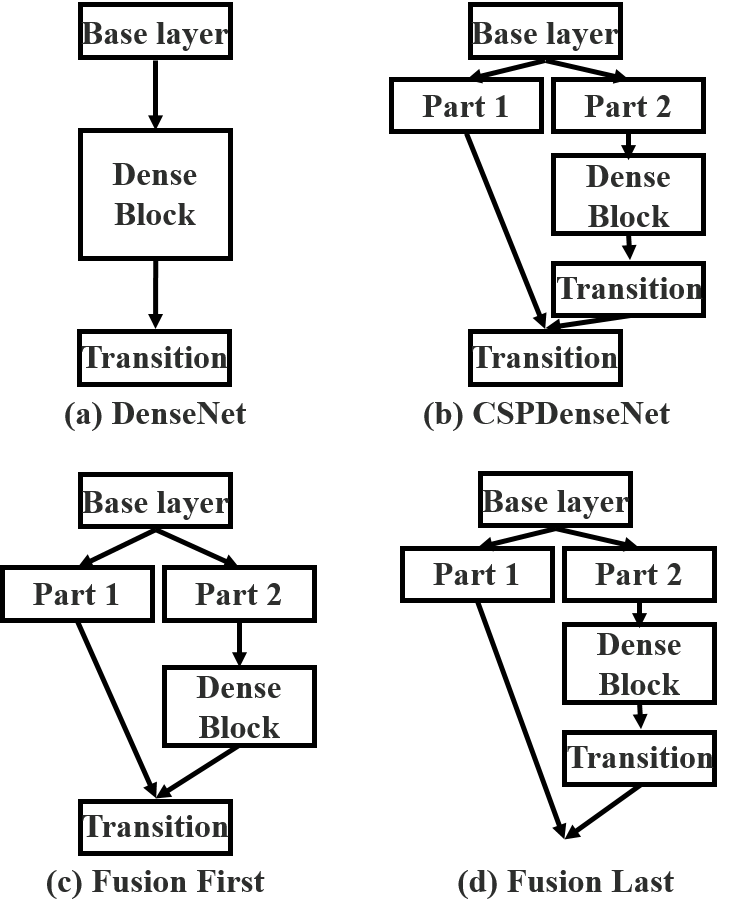}
	\end{center}
	\caption{Different kind of feature fusion strategies.  (a) single path DenseNet, (b) proposed CSPDenseNet: transition $\rightarrow$ concatenation $\rightarrow$ transition, (c) concatenation $\rightarrow$ transition, and (d) transition $\rightarrow$ concatenation.  }
	\label{fig:fusion}
\end{figure}

\begin{figure}[h]
	\begin{center}
		\includegraphics[width=0.6\linewidth]{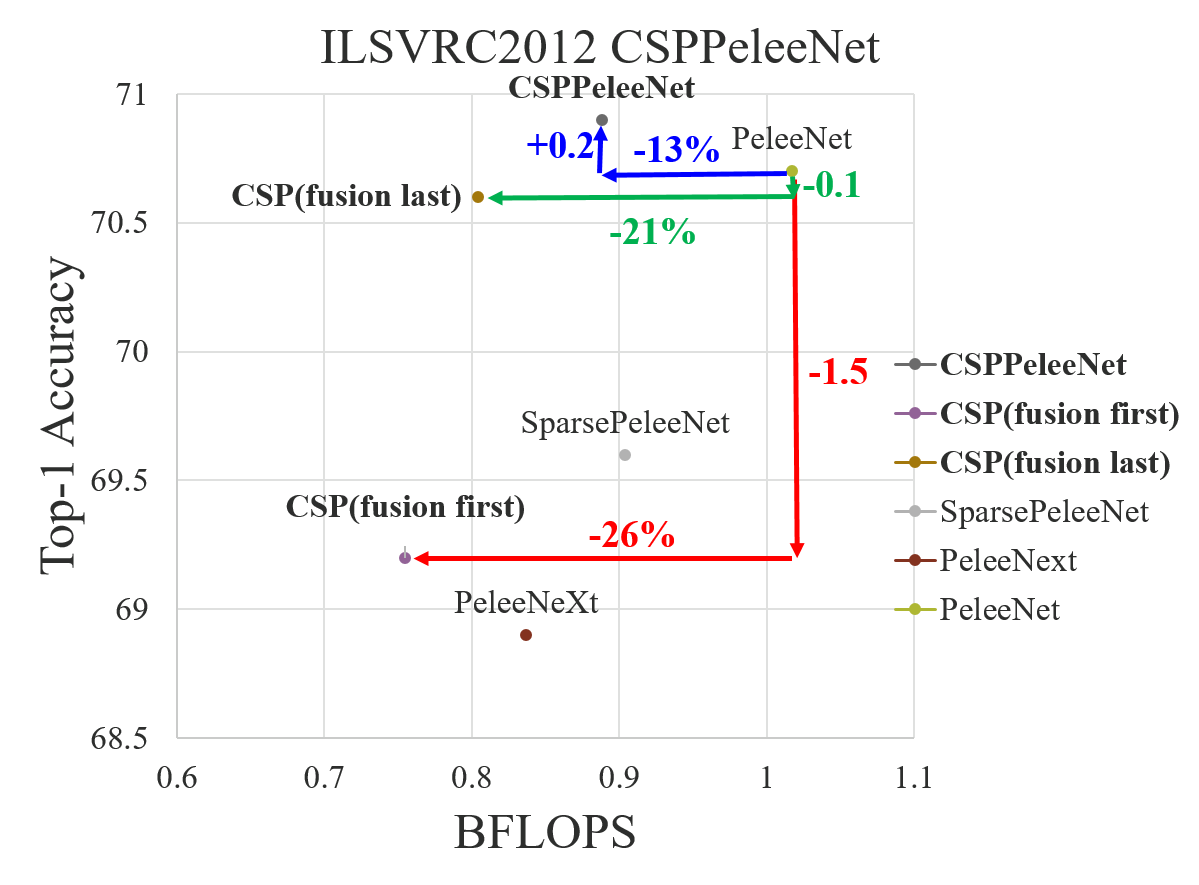}
	\end{center}
	\caption{Effect of truncating gradient flow for maximizing difference of gradient combination.  }
	\label{fig:fusion_ab}
\end{figure}

{\bf Partial Transition Layer.} The purpose of designing partial transition layers is to maximize the difference of gradient combination.  The partial transition layer is a hierarchical feature fusion mechanism, which uses the strategy of truncating the gradient flow to prevent distinct layers from learning duplicate gradient information.  Here we design two variations of CSPDenseNet to show how this sort of gradient flow truncating affects the learning ability of a network.  \ref{fig:fusion} (c) and \ref{fig:fusion} (d) show two different fusion strategies.  CSP (fusion first) means to concatenate the feature maps generated by two parts, and then do transition operation.  If this strategy is adopted, a large amount of gradient information will be reused.  As to the CSP (fusion last) strategy, the output from the dense block will go through the transition layer and then do concatenation with the feature map coming from part 1.  If one goes with the CSP (fusion last) strategy, the gradient information will not be reused since the gradient flow is truncated.  If we use the four architectures shown in \ref{fig:fusion} to perform image classification, the corresponding results are shown in Figure \ref{fig:fusion_ab}.  It can be seen that if one adopts the CSP (fusion last) strategy to perform image classification, the computation cost is significantly dropped, but the top-1 accuracy only drop 0.1\%.  On the other hand, the CSP (fusion first) strategy does help the significant drop in computation cost, but the top-1 accuracy significantly drops 1.5\%.  By using the split and merge strategy across stages, we are able to effectively reduce the possibility of duplication during the information integration process.  From the results shown in Figure \ref{fig:fusion_ab}, it is obvious that if one can effectively reduce the repeated gradient information, the learning ability of a network will be greatly improved.

\begin{figure}[h]
	\begin{center}
		\includegraphics[width=0.5\linewidth]{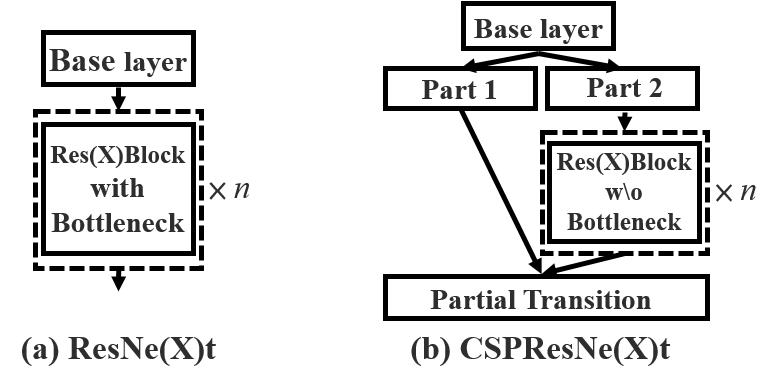}
	\end{center}
	\caption{Applying CSPNet to ResNe(X)t.  }
	\label{fig:cspresnext}
\end{figure}

{\bf Apply CSPNet to Other Architectures.} CSPNet can be also easily applied to ResNet and ResNeXt, the architectures are shown in Figure \ref{fig:cspresnext}.  Since only half of the feature channels are going through Res(X)Blocks, there is no need to introduce the bottleneck layer anymore.  This makes the theoretical lower bound of the Memory Access Cost (MAC) when the FLoating-point OPerations (FLOPs) is fixed.

\subsection{Exact Fusion Model}

\begin{figure*}[t]
	\begin{center}
		\includegraphics[width=0.95\linewidth]{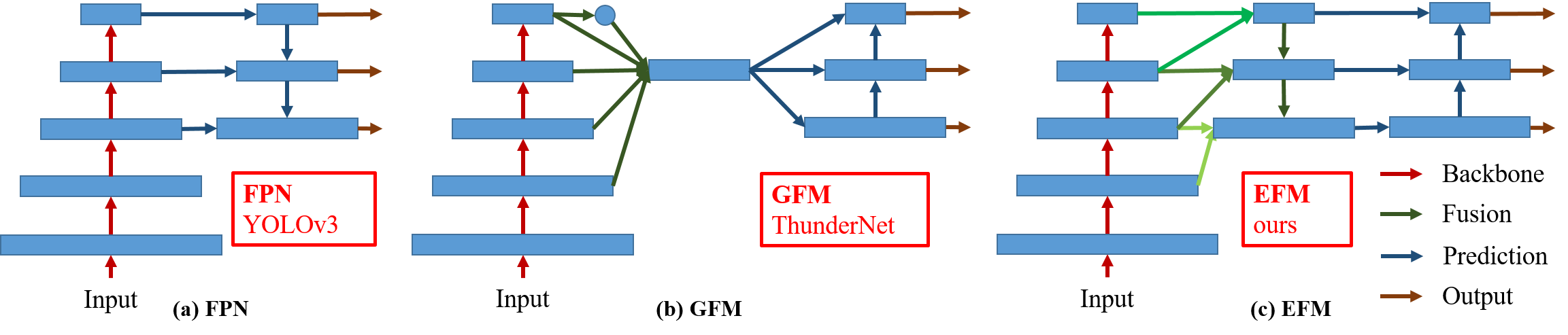}
	\end{center}
	\caption{Different feature pyramid fusion strategies.  (a) Feature Pyramid Network (FPN): fuse features from current scale and previous scale.  (b) Global Fusion Model (GFM): fuse features of all scales.  (c) Exact Fusion Model (EFM): fuse features depand on anchor size.   }
	\label{fig:detector}
\end{figure*}


{\bf Looking Exactly to predict perfectly.} We propose EFM that captures an appropriate Field of View (FoV) for each anchor, which enhances the accuracy of the one-stage object detector.  For segmentation tasks, since pixel-level labels usually do not contain global information, it is usually more preferable to consider larger patches for better information retrieval \cite{liu2015parsenet}. However, for tasks like image classification and object detection, some critical information can be obscure when observed from image-level and bounding box-level labels. Li \textit{et al.} \cite{li2018tell} found that CNN can be often distracted when it learns from image-level labels and concluded that it is one of the main reasons that two-stage object detectors outperform one-stage object detectors.

{\bf Aggregate Feature Pyramid.} The proposed EFM is able to better aggregate the initial feature pyramid. The EFM is based on YOLOv3 \cite{redmon2018yolov3}, which assigns exactly one bounding-box prior to each ground truth object. Each ground truth bounding box corresponds to one anchor box that surpasses the threshold IoU. If the size of an anchor box is equivalent to the FoV of the grid cell, then for the grid cells of the $s^{th}$ scale, the corresponding bounding box will be lower bounded by the $(s-1)^{th}$ scale and upper bounded by the $(s+1)^{th}$ scale. Therefore, the EFM assembles features from the three scales.

{\bf Balance Computation.} Since the concatenated feature maps from the feature pyramid are enormous, it introduces a great amount of memory and computation cost. To alleviate the problem, we incorporate the Maxout technique to compress the feature maps.

\section{Experiments}

We will use ImageNet's image classification dataset \cite{deng2009imagenet} used in ILSVRC 2012 to validate our proposed CSPNet.  Besides, we also use the MS COCO object detection dataset \cite{lin2014microsoft} to verify the proposed EFM.  Details of the proposed architectures will be elaborated in the appendix.

\subsection{Implementation Details}

{\bf ImageNet.} In ImageNet image classification experiments, all hyper-parameters such as training steps, learning rate schedule, optimizer, data augmentation, etc., we all follow the settings defined in Redmon \textit{et al.} \cite{redmon2018yolov3}.  For ResNet-based models and ResNeXt-based models, we set 8000,000 training steps.  As to DenseNet-based models, we set 1,600,000 training steps.  We set the initial learning rate 0.1 and adopt the polynomial decay learning rate scheduling strategy.  The momentum and weight decay are respectively set as 0.9 and 0.005.  All architectures use a single GPU to train universally in the batch size of 128.  Finally, we use the validation set of ILSVRC 2012 to validate our method.  

{\bf MS COCO.} In MS COCO object detection experiments, all hyper-parameters also follow the settings defined in Redmon \textit{et al.} \cite{redmon2018yolov3}.  Altogether we did 500,000 training steps.  We adopt the step decay learning rate scheduling strategy and multiply with a factor 0.1 at the 400,000 steps and the 450,000 steps, respectively.  The momentum and weight decay are respectively set as 0.9 and 0.0005.  All architectures use a single GPU to execute multi-scale training in the batch size of 64.  Finally, the COCO test-dev set is adopted to verify our method.

\subsection{Ablation Experiments}

{\bf Ablation study of CSPNet on ImageNet.} In the ablation experiments conducted on the CSPNet, we adopt PeleeNet \cite{wang2018pelee} as the baseline, and the ImageNet is used to verify the performance of the CSPNet.  We use different partial ratios $\gamma$ and the different feature fusion strategies for ablation study.  Table \ref{table:imagenet_ab} shows the results of ablation study on CSPNet.  In Table \ref{table:imagenet_ab}, SPeleeNet and PeleeNeXt are, respectively, the architectures that introduce sparse connection and group convolution to PeleeNet.  As to CSP (fusion first) and CSP (fusion last), they are the two strategies proposed to validate the benefits of a partial transition. 

\begin{table}[h]
	\centering
	\begin{threeparttable}[h]
		\footnotesize
		\caption{Ablation study of CSPNet on ImageNet.}
		\label{table:imagenet_ab}
		\setlength\tabcolsep{2.0pt}
		\begin{tabular}{lccccccc}
			\toprule
			Model & $\gamma$ & \begin{tabular}{@{}c@{}}two-way \\ dense\end{tabular} & \begin{tabular}{@{}c@{}}partial \\ dense\end{tabular} & trans. & \begin{tabular}{@{}c@{}}partial \\ trans.\end{tabular} & Top-1 & BFLOPs \\			
			\midrule
			PeleeNet \cite{wang2018pelee} & - & $\checkmark$ &  & $\checkmark$ &  & 70.7 & 1.017 \\	
			SPeleeNet & - & $\checkmark$(sparse) &  & $\checkmark$ &  & 69.6 & 0.904 \\	
			PeleeNeXt & - & $\checkmark$(group) &  & $\checkmark$ &  & 68.9 & 0.837 \\	
			\midrule
			& 0.75 &  &  &  &  & 68.4 & 0.649 \\	
			\textbf{CSP (fusion first)} & 0.5 &  & $\checkmark$ & $\checkmark$ &  & 69.2 & 0.755 \\	
			& 0.25 &  &  &  &  & 70.0 & 0.861 \\
			\midrule	
			& 0.75 &  &  &  &  & 69.2 & 0.716 \\	
			\textbf{CSP (fusion last)} & 0.5 &  & $\checkmark$ &  & $\checkmark$ & 70.6 & 0.804 \\	
			& 0.25 &  &  &  &  & \textbf{70.8} & \textbf{0.902} \\
			\midrule	
			& 0.75 &  &  &  &  & 70.4 & 0.800 \\	
			\textbf{CSPPeleeNet} & 0.5 &  & $\checkmark$ &  & $\checkmark$ & \textbf{70.9} & \textbf{0.888} \\	
			& 0.25 &  &  &  &  & \textbf{71.5} & \textbf{0.986} \\	
			\bottomrule
		\end{tabular}
	\end{threeparttable}
\end{table}

From the experimental results, if one only uses the CSP (fusion first) strategy on the cross-stage partial dense block, the performance can be slightly better than SPeleeNet and PeleeNeXt.  However, the partial transition layer designed to reduce the learning of redundant information can achieve very good performance.  For example, when the computation is cut down by 21\%, the accuracy only degrades by 0.1\%.  One thing to be noted is that when $\gamma = 0.25$, the computation is cut down by 11\%, but the accuracy is increased by 0.1\%.  Compared to the baseline PeleeNet, the proposed CSPPeleeNet achieves the best performance, it can cut down 13\% computation, but at the same time upgrade the accuracy by 0.2\%.  If we adjust the partial ratio to $\gamma = 0.25$, we are able to upgrade the accuracy by 0.8\% and at the same time cut down 3\% computation.

{\bf Ablation study of EFM on MS COCO.} Next, we shall conduct an ablation study of EFM based on the MS COCO dataset.  In this series of experiments, we compare three different feature fusion strategies shown in Figure \ref{fig:detector}.  We choose two state-of-the-art lightweight models, PRN \cite{wang2019enriching} and ThunderNet \cite{qin2019thundernet}, to make comparison.  PRN is the feature pyramid architecture used for comparison, and the ThunderNet with Context Enhancement Module (CEM) and Spatial Attention Module (SAM) are the global fusion architecture used for comparison.  We design a Global Fusion Model (GFM) to compare with the proposed EFM.  Moreover, GIoU \cite{rezatofighi2019generalized}, SPP, and SAM are also applied to EFM to conduct an ablation study.  All experiment results listed in Table \ref{table:coco_ab} adopt CSPPeleeNet as the backbone.

\begin{table}[h]
	\centering
	\begin{threeparttable}[h]
		\footnotesize
		\caption{Ablation study of EFM on MS COCO.}
		\label{table:coco_ab}
		\setlength\tabcolsep{2.0pt}
		\begin{tabular}{lcccccccc}
			\toprule
			Head & \begin{tabular}{@{}c@{}}global \\ fusion\end{tabular} & \begin{tabular}{@{}c@{}}exact \\ fusion\end{tabular} & atten. & BFLOPs & FPS & AP & AP$_{50}$ & AP$_{75}$ \\			
			\midrule
			PRN \cite{wang2019enriching} &  &  &  & 3.590 & 169 & 23.1 & 44.5 & 22.0 \\
			PRN (swish) &  &  &  & 3.590 & 161 & 24.1 & 45.8 & 23.3 \\
			PRN-3l \cite{wang2019enriching} &  &  &  & 4.586 & 151 & 23.7 & 46.0 & 22.2 \\		
			\midrule
			CEM \cite{qin2019thundernet} & $\checkmark$ &  &  & 4.049 & 148 & 23.8 & 45.4 & 22.6 \\
			CEM (SAM) \cite{qin2019thundernet} & $\checkmark$ &  & $\checkmark$ & 4.165 & 144 & 24.1 & 46.0 & 23.1 \\
			GFM & $\checkmark$ &  &  & 4.605 & 134 & 24.3 & 46.2 & 23.3 \\		
			\midrule
			\textbf{EFM} &  & $\checkmark$ &  & 4.868 & 132 & 26.4 & 48.6 & 26.3 \\
			\textbf{EFM (GIoU \cite{rezatofighi2019generalized})} &  & $\checkmark$ &  & 4.868 & 132 & \textbf{27.1} & 45.9 & \textbf{28.2} \\
			\textbf{EFM (SAM)} &  & $\checkmark$ & $\checkmark$ & 5.068 & 129 & 26.8 & \textbf{49.0} & 26.7 \\
			\textbf{EFM (SPP)} &  & $\checkmark$ &  & 4.863 & 128 & 26.2 & 48.5 & 25.7 \\
			\bottomrule
		\end{tabular}
	\end{threeparttable}
\end{table}

As reflected in the experiment results, the proposed EFM is 2 fps slower than GFM, but its AP and AP$_{50}$ are significantly upgraded by 2.1\% and 2.4\%, respectively.  Although the introduction of GIoU can upgrade AP by 0.7\%, the AP$_{50}$ is, however, significantly degraded by 2.7\%.  However, for edge computing, what really matters is the number and locations of the objects rather than their coordinates.  Therefore, we will not use GIoU training in the subsequent models.  The attention mechanism used by SAM can get a better frame rate and AP compared with SPP's increase of FoV mechanism, so we use EFM (SAM) as the final architecture.  In addition, although the CSPPeleeNet with swish activation can improve AP by 1\%, its operation requires a lookup table on the hardware design to accelerate, we finally also abandoned the swish activation function.

\subsection{ImageNet Image Classification}

We apply the proposed CSPNet to ResNet-10 \cite{he2016deep}, ResNeXt-50 \cite{xie2017aggregated}, PeleeNet \cite{wang2018pelee}, and DenseNet-201-Elastic \cite{wang2019elastic} and compare with state-of-the-art methods.  The experimental results are shown in Table \ref{table:imagenet}.

\begin{table}[h]
	\centering
	\begin{threeparttable}[h]
		\footnotesize
		\caption{Compare with state-of-the-art methods on ImageNet.}
		\label{table:imagenet}
		\setlength\tabcolsep{2.0pt}
		\begin{tabular}{lcccc}
			\toprule
			Model & \#Parameter & BFLOPs & Top-1 & Top-5 \\			
			\midrule	
			\midrule
			PeleeNet \cite{wang2018pelee} & 2.79M & 1.017 & 70.7\% & 90.0\% \\	
			PeleeNet-swish & 2.79M & 1.017 & 71.5\% & 90.7\% \\
			SEPeleeNet-swish & 2.81M & 1.017 & 72.1\% & \textbf{91.0\%} \\
			\textbf{CSPPeleeNet} & 2.83M & 0.888 \textbf{\textcolor{blue}{(-13\%)}} & 70.9\%  & 90.2\% \\	
			\textbf{CSPPeleeNet-swish} & 2.83M & 0.888 \textbf{\textcolor{blue}{(-13\%)}} & 71.7\% & 90.8\% \\
			\textbf{SECSPPeleeNet-swish} & 2.85M & 0.888 \textbf{\textcolor{blue}{(-13\%)}} & \textbf{72.4\%} & \textbf{91.0\%} \\
			SparsePeleeNet \cite{zhu2018sparsely} & 2.39M & 0.904 & 69.6\% & 89.3\% \\
			\midrule	
			EfficientNet-B0* \cite{tan2019efficientnet} & 4.81M  & 0.915 & \textbf{71.3\%} & \textbf{90.4\%} \\
			EfficientNet-B0** \cite{tan2019efficientnet} & -  & - & 70.0\% & 88.9\% \\
			\midrule
			Darknet Reference \cite{darknet13} & 7.31M & 0.96 & 61.1\% & 83.0\% \\
			\textbf{CSPDenseNet Reference} & 3.48M & 0.886 & 65.7\% & 86.6\% \\
			\textbf{CSPPeleeNet Reference} & 4.10M & 1.103 & \textbf{68.9\%} & \textbf{88.7\%} \\
			\midrule
			ResNet-10 \cite{he2016deep} & 5.24M & 2.273 & 63.5\% & 85.0\% \\
			\textbf{CSPResNet-10} & 2.73M & 1.905 \textbf{\textcolor{blue}{(-16\%)}} & \textbf{65.3\%} & \textbf{86.5\%} \\
			\midrule
			ResNeXt-50 \cite{xie2017aggregated} & 22.19M & 10.11 & 77.8\%  & \textbf{94.2\%} \\
			\textbf{CSPResNeXt-50} & 20.50M & 7.93 \textbf{\textcolor{blue}{(-22\%)}} & \textbf{77.9\%} & 94.0\% \\
			HarDNet-138s \cite{chao2019hardnet} & 35.5M & 13.4 & 77.8\% & - \\
			DenseNet-264-32 \cite{huang2017densely} & 27.21M & 11.03 & 77.8\% & 93.9\% \\
			ResNet-152 \cite{he2016deep} & 60.2M & 22.6 & 77.8\% & 93.6\% \\
			\midrule
			DenseNet-201-Elastic \cite{wang2019elastic} & 19.48M & 8.77 &  \textbf{77.9\%} &  \textbf{94.0\%} \\
			\textbf{CSPDenseNet-201-Elastic} & 20.17M & 7.13 \textbf{\textcolor{blue}{(-19\%)}} & \textbf{77.9\%} & \textbf{94.0\%} \\
			\midrule
			Res2Net-50 (10 crop) \cite{gao2019res2net} & 25.29M & 8.4$\times$10 & 78.0\% & 93.8\% \\
			Res2NeXt-50 (10 crop) \cite{gao2019res2net} & 24.27M & 8.4$\times$10 & \textbf{78.2\%} & 93.9\% \\
			\textbf{CSPResNeXt-50} (10 crop) & 20.50M & 7.9$\times$10 & \textbf{78.2\%} & \textbf{94.3\%} \\
			\bottomrule
		\end{tabular}
		\begin{tablenotes}[flushleft]
			\footnotesize
			\item[1] EfficientNet* is implemented by Darknet framework.
			\item[2] EfficientNet** is trained by official code with batch size 256.
			\item[3] Swish activation function is presented by \cite{elfwing2018sigmoid, ramachandran2017searching}.
			\item[4] Squeeze-and-excitation (SE) network is presented by \cite{hu2018squeeze}.
		\end{tablenotes}
	\end{threeparttable}
\end{table}

It is confirmed by experimental results that no matter it is ResNet-based models, ResNeXt-based models, or DenseNet-based models, when the concept of CSPNet is introduced, the computational load is reduced at least by 10\% and the accuracy is either remain unchanged or upgraded.  Introducing the concept of CSPNet is especially useful for the improvement of lightweight models.  For example, compared to ResNet-10, CSPResNet-10 can improve accuracy by 1.8\%.  As to PeleeNet and DenseNet-201-Elastic, CSPPeleeNet and CSPDenseNet-201-Elastic can respectively cut down 13\% and 19\% computation, and either upgrade a little bit or maintain the accuracy.  As to the case of ResNeXt-50, CSPResNeXt-50 can cut down 22\% computation and upgrade top-1 accuracy to 77.9\%.

\begin{table*}[t]
	\centering
	\begin{threeparttable}[t]
		\footnotesize
		\caption{Compare with state-of-the-art methods on MSCOCO Object Detection.}
		\label{table:mscoco}
		\setlength\tabcolsep{3.5pt}
		\begin{tabular}{llcccccccccc}
			\toprule
			Method & Backbone & Size & FPS & BFLOPs & \#Parameter & AP & AP$_{50}$ & AP$_{75}$ & AP$_{S}$ & AP$_{M}$ & AP$_{L}$ \\			
			\midrule	
			YOLOv3 \cite{redmon2018yolov3} & DarkNet53 \cite{redmon2018yolov3} & 608$\times$608 & 30 & 140.7 & 62.3M & 33.0 & 57.9 & 34.4 & 18.3 & 25.4 & 41.9 \\
			YOLOv3 (SPP) \cite{redmon2018yolov3} & DarkNet53 \cite{redmon2018yolov3} & 608$\times$608 & 30 & 141.5 & 62.9M & 36.2 & \textbf{60.6} & 38.2 & 20.6 & 37.4 & 46.1 \\	
			LRF \cite{wang2019learning} & ResNet101 \cite{he2016deep} & 512$\times$512 & 31 & - & - & 37.3 & 58.5 & 39.7 & 19.7 & \textbf{42.8} & 50.1 \\
			SSD \cite{liu2016ssd} & HarDNet85 \cite{chao2019hardnet} & 512$\times$512 & 32 & - & - & 35.1 & 54.8 & 37.6 & 15.0 & 38.9 & \textbf{51.5} \\	
			M2Det \cite{zhao2019m2det} & VGG16 \cite{simonyan2014very} & 320$\times$320 & 33 & - & - & 33.5 & 52.4 & 35.6 & 14.4 & 37.6 & 47.6 \\
			PFPNet (R) \cite{kim2018parallel} & VGG16 \cite{simonyan2014very} & 320$\times$320 & 33 & - & - & 31.8 & 52.9 & 33.6 & 12.0 & 35.5 & 46.1 \\	
			DAFS \cite{li2019dynamic} & VGG16 \cite{simonyan2014very} & 512$\times$512 & 35 & - & - & 33.8 & 52.9 & 36.9 & 14.6 & 37.0 & 47.7 \\
			RFBNet \cite{liu2018receptive} & VGG16 \cite{simonyan2014very} & 512$\times$512 & 35 & - & - & 33.8 & 54.2 & 35.9 & 16.2 & 37.1 & 47.4 \\
			PANet (SPP) \cite{liu2018path} & \textbf{CSPResNeXt50} &  608$\times$608 & 35 & 100.6 & 56.9M & \textbf{38.4} & \textbf{60.6} & \textbf{41.6} & \textbf{22.1} & 41.8 & 47.6 \\
			SSD \cite{liu2016ssd} & HarDNet68 \cite{chao2019hardnet} & 512$\times$512 & 38 & - & - & 31.7 & 51.0 & 33.8 & 12.5 & 35.1 & 47.9 \\	
			LRF \cite{wang2019learning} & VGG16 \cite{simonyan2014very} & 512$\times$512 & 38 & - & - & 36.2 & 56.6 & 38.7 & 19.0 & 39.9 & 48.8 \\
			PFPNet (S) \cite{kim2018parallel} & VGG16 \cite{simonyan2014very} & 300$\times$300 & 39 & - & - & 29.6 & 49.6 & 31.1 & 10.6 & 32.0 & 44.9 \\	
			RefineDet \cite{zhang2018single} & VGG16 \cite{simonyan2014very} & 320$\times$320 & 40 & - & - & 29.4 & 49.2 & 31.3 & 10.0 & 32.0 & 44.4 \\	
			SSD \cite{liu2016ssd} & VGG16 \cite{simonyan2014very} & 300$\times$300 & 44 & 70.4 & 34.3M & 25.7 & 43.9 & 26.2 & 6.9 & 27.7 & 42.6 \\	
			PANet (SPP) \cite{liu2018path} & \textbf{CSPResNeXt50} & 512$\times$512 & 44 & 71.3 & 56.9M & 38.0 & 60.0 & 40.8 & 19.7 & 41.4 & 49.9 \\
			CenterNet \cite{zhou2019objects} & ResNet101 \cite{he2016deep} & 512$\times$512 & 45 & - & - & 34.6 & 53.0 & 36.9 &  &  &  \\	
			YOLOv3 \cite{redmon2018yolov3} & DarkNet53 \cite{redmon2018yolov3} & 416$\times$416 & 46 & 65.9 & 62.3M & 31.0 & 55.3 & 32.3 & 15.2 & 33.2 & 42.8 \\	
			PANet (SPP) \cite{liu2018path} & \textbf{CSPResNeXt50} & 416$\times$416 & 53 & 47.1 & 56.9M & 36.6 & 58.1 & 39.0 & 16.2 & 39.5 & 50.9 \\
			TTFNet \cite{liu2019training} & DarkNet53 \cite{redmon2018yolov3} & 512$\times$512 & 54 & - & - & 35.1 & 52.5 & 37.8 & 17.0 & 38.5 & 49.5 \\	
			YOLOv3 \cite{redmon2018yolov3} & DarkNet53 \cite{redmon2018yolov3} & 320$\times$320 & 56 & 39.0 & 62.3M & 28.2 & 51.5 & 29.7 & 11.9 & 30.6 & 43.4 \\	
			PANet (SPP) \cite{liu2018path} & \textbf{CSPResNeXt50} & 320$\times$320 & 58 & 27.9 & 56.9M & 33.4 & 54.0 & 35.1 & 11.8 & 35.3 & 50.9 \\				
			\midrule
			Pelee \cite{wang2018pelee} & PeleeNet \cite{wang2018pelee} & 304$\times$304 & 106 & 2.58 & 5.98M & 22.4 & 38.3 & 22.9 &  &  &  \\
			\textbf{EFM (SAM)} & \textbf{CSPPeleeNet} & 512$\times$512 & 109 & 7.68 & 4.31M & 27.6 & \textbf{50.4} & 27.7 & \textbf{12.4} & \textbf{30.1} & 36.2 \\
			TTFNet \cite{liu2019training} & ResNet18 \cite{he2016deep} & 512$\times$512 & 112 & - & - & \textbf{28.1} & 43.8 & \textbf{30.2} & 11.8 & 29.5 & \textbf{41.5} \\
			CenterNet \cite{zhou2019objects} & ResNet18 \cite{he2016deep} & 512$\times$512 & 129 & - & - & \textbf{28.1} & 44.9 & 29.6 &  &  &  \\
			\textbf{EFM (SAM)} & \textbf{CSPPeleeNet} & 416$\times$416 & 129 & 5.07 & 4.31M & 26.8 & 49.0 & 26.7 & 9.8 & 28.2 & 38.8 \\
			PRN \cite{wang2019enriching} & PeleeNet \cite{wang2018pelee} & 416$\times$416 & 145 & 4.04 & 3.16M & 23.3 & 45.0 & 22.0 & 6.7 & 24.8 & 35.1 \\
			\midrule
			\textbf{EFM (SAM)} \cite{wang2019enriching} & \textbf{CSPPeleeNet Ref.}  & 320$\times$320 & 205 & 3.43 & 5.67M & 23.5 & 44.6 & 22.7 & 7.1 & 23.6 & 36.1 \\	
			ThunderNet \cite{qin2019thundernet} & SNet535 \cite{qin2019thundernet} & 320$\times$320 & 214 & 2.60 & - & 28.0 & 46.2 & 29.5 &  &  &  \\
			\textbf{EFM (SAM)} \cite{wang2019enriching} & \textbf{CSPDenseNet Ref.}  & 320$\times$320 & 235 & 2.89 & 5.05M & 21.7 & 42.2 & 20.6 & 6.3 & 21.3 & 33.3 \\	
			ThunderNet \cite{qin2019thundernet} & SNet146 \cite{qin2019thundernet} & 320$\times$320 & 248 & 0.95 & - & 23.6 & 40.2 & 24.5 &  &  &  \\
			ThunderNet \cite{qin2019thundernet} & SNet49 \cite{qin2019thundernet} & 320$\times$320 & 267 & 0.52 & - & 19.1 & 33.7 & 19.6 &  &  & \\	
			PRN (3l) \cite{wang2019enriching} & \textbf{CSPPeleeNet Ref.}  & 320$\times$320 & 267 & 3.15 & 4.79M & 19.4 & 40.0 & 17.0 & 5.8 & 18.8 & 31.1 \\				
			\midrule
			PRN \cite{wang2019enriching} & \textbf{CSPPeleeNet Ref.}  & 320$\times$320 & 306 & 2.56 & 4.59M & 18.8 & 38.5 & 16.6 & 5.0 & 18.9 & 31.4 \\			
			YOLOv3 (tiny) \cite{redmon2018yolov3} & DarkNet Ref. \cite{redmon2018yolov3} & 416$\times$416 & 330 & 5.57 & 8.86M &  & 33.1 &  &  &  &  \\
			PRN \cite{wang2019enriching} & \textbf{CSPDenseNet Ref.}  & 320$\times$320 & 387 & 2.01 & 3.97M & 16.8 & 35.4 & 14.3 & 4.4 & 16.6 & 28.5 \\			
			PRN \cite{wang2019enriching} & DarkNet Ref. \cite{redmon2018yolov3} & 416$\times$416 & 400 & 3.47 & 4.96M &  & 33.1 &  &  &  &  \\
			PRN \cite{wang2019enriching} & \textbf{CSPDenseNetb Ref.}  & 320$\times$320 & 400 & 1.59 & 1.87M & 15.3 & 34.2 & 12.0 & 3.6 & 16.1 & 23.4 \\				
			\midrule
			\bottomrule
		\end{tabular}
		\begin{tablenotes}[flushleft]
			\footnotesize
			\item[1] The table is separated into four parts, $<$100 fps, 100$\sim$200 fps, 200$\sim$300 fps, and $>$300 fps.
			\item[2] We mainly focus on \textbf{FPS} and \textbf{AP$_{50}$} since almost all applications need fast inference to locate and count objects.
			\item[3] Inference speed are tested on \textbf{GTX 1080ti} with batch size equals to 1 if possible, and our models are tested using Darknet \cite{alexeyab84_darknet}.
			\item[4] All results are obtained by \textbf{COCO test-dev set} except for TTFNet \cite{liu2019training} models which are verified on minval5k set.
		\end{tablenotes}
	\end{threeparttable}
\end{table*}

If compared with the state-of-the-art lightweight model -- EfficientNet-B0, although it can achieve 76.8\% accuracy when the batch size is 2048, when the experiment environment is the same as ours, that is, only one GPU is used, EfficientNet-B0 can only reach 70.0\% accuracy.  In fact, the swish activation function and SE block used by EfficientNet-B0 are not efficient on the mobile GPU.  A similar analysis has been conducted during the development of EfficientNet-EdgeTPU.  Here, for demonstrating the learning ability of CSPNet, we introduce swish and SE into CSPPeleeNet and then make a comparison with EfficientNet-B0*.  In this experiment, SECSPPeleeNet-swish cut down computation by 3\% and upgrade 1.1\% top-1 accuracy.

Proposed CSPResNeXt-50 is compared with ResNeXt-50 \cite{xie2017aggregated}, ResNet-152 \cite{he2016deep}, DenseNet-264 \cite{huang2017densely}, and HarDNet-138s \cite{chao2019hardnet}, regardless of parameter quantity, amount of computation, and top-1 accuracy, CSPResNeXt-50 all achieve the best result.  As to the 10-crop test, CSPResNeXt-50 also outperforms Res2Net-50 \cite{gao2019res2net} and Res2NeXt-50 \cite{gao2019res2net}.

\subsection{MS COCO Object Detection}

In the task of object detection, we aim at three targeted scenarios: (1) real-time on GPU: we adopt CSPResNeXt50 with PANet (SPP) \cite{liu2018path}; (2) real-time on mobile GPU: we adopt CSPPeleeNet, CSPPeleeNet Reference, and CSPDenseNet Reference with the proposed EFM (SAM); and (3) real-time on CPU: we adopt CSPPeleeNet Reference and CSPDenseNet Reference with PRN \cite{wang2019enriching}.  The comparisons between the above models and the state-of-the-art methods are listed in Table \ref{table:mscoco}.  As to the analysis on the inference speed of CPU and mobile GPU will be detailed in the next subsection.

If compared to object detectors running at 30$\sim$100 fps, CSPResNeXt50 with PANet (SPP) achieves the best performance in AP, AP$_{50}$ and AP$_{75}$.  They receive, respectively, 38.4\%, 60.6\%, and 41.6\% detection rates.  If compared to state-of-the-art LRF \cite{wang2019learning} under the input image size 512$\times$512, CSPResNeXt50 with PANet (SPP) outperforms ResNet101 with LRF by 0.7\% AP, 1.5\% AP$_{50}$ and 1.1\% AP$_{75}$.  If compared to object detectors running at 100$\sim$200 fps, CSPPeleeNet with EFM (SAM) boosts 12.1\% AP$_{50}$ at the same speed as Pelee \cite{wang2018pelee} and increases 4.1\% \cite{wang2018pelee} at the same speed as CenterNet \cite{zhou2019objects}.

If compared to very fast object detectors such as ThunderNet \cite{qin2019thundernet}, YOLOv3-tiny \cite{redmon2018yolov3}, and YOLOv3-tiny-PRN \cite{wang2019enriching}, the proposed CSPDenseNetb Reference with PRN is the fastest.  It can reach 400 fps frame rate, i.e., 133 fps faster than ThunderNet with SNet49.  Besides, it gets 0.5\% higher on AP$_{50}$.  If compared to ThunderNet146, CSPPeleeNet Reference with PRN (3l) increases the frame rate by 19 fps while maintaining the same level of AP$_{50}$.

\subsection{Analysis}

\begin{figure}[h]
	\begin{center}
		\includegraphics[width=0.95\linewidth]{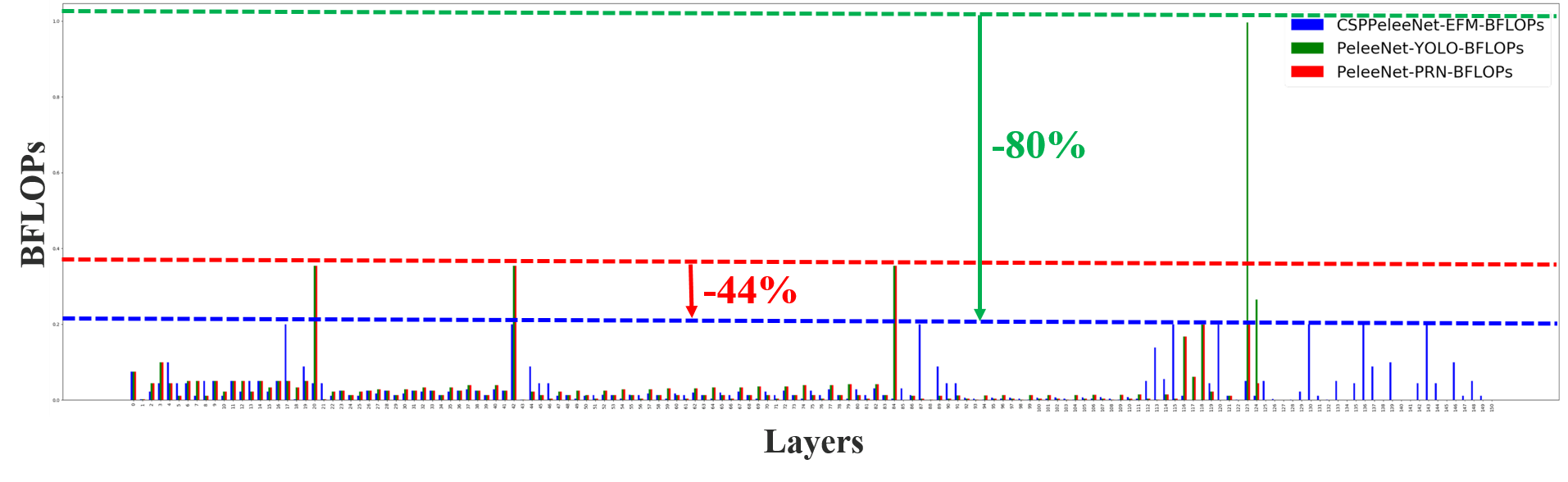}
	\end{center}
	\caption{Computational bottleneck of PeleeNet-YOLO, PeleeNet-PRN and CSPPeleeNet-EFM. }
	\label{fig:bottleneck}
\end{figure}

{\bf Computational Bottleneck.} Figure \ref{fig:bottleneck} shows the BLOPS of each layer of PeleeNet-YOLO, PeleeNet-PRN and proposed CSPPeleeNet-EFM.  From Figure \ref{fig:bottleneck}, it is obvious that the computational bottleneck of PeleeNet-YOLO occurs when the head integrates the feature pyramid.  The computational bottleneck of PeleeNet-PRN occurs on the transition layers of the PeleeNet backbone.  As to the proposed CSPPeleeNet-EFM, it can balance the overall computational bottleneck, which reduces the PeleeNet backbone 44\% computational bottleneck and reduces PeleeNet-YOLO 80\% computational bottleneck.  Therefore, we can say that the proposed CSPNet can provide hardware with a higher utilization rate.

\vspace*{1mm}

\begin{figure}[h]
	\begin{center}
		\includegraphics[width=0.8\linewidth]{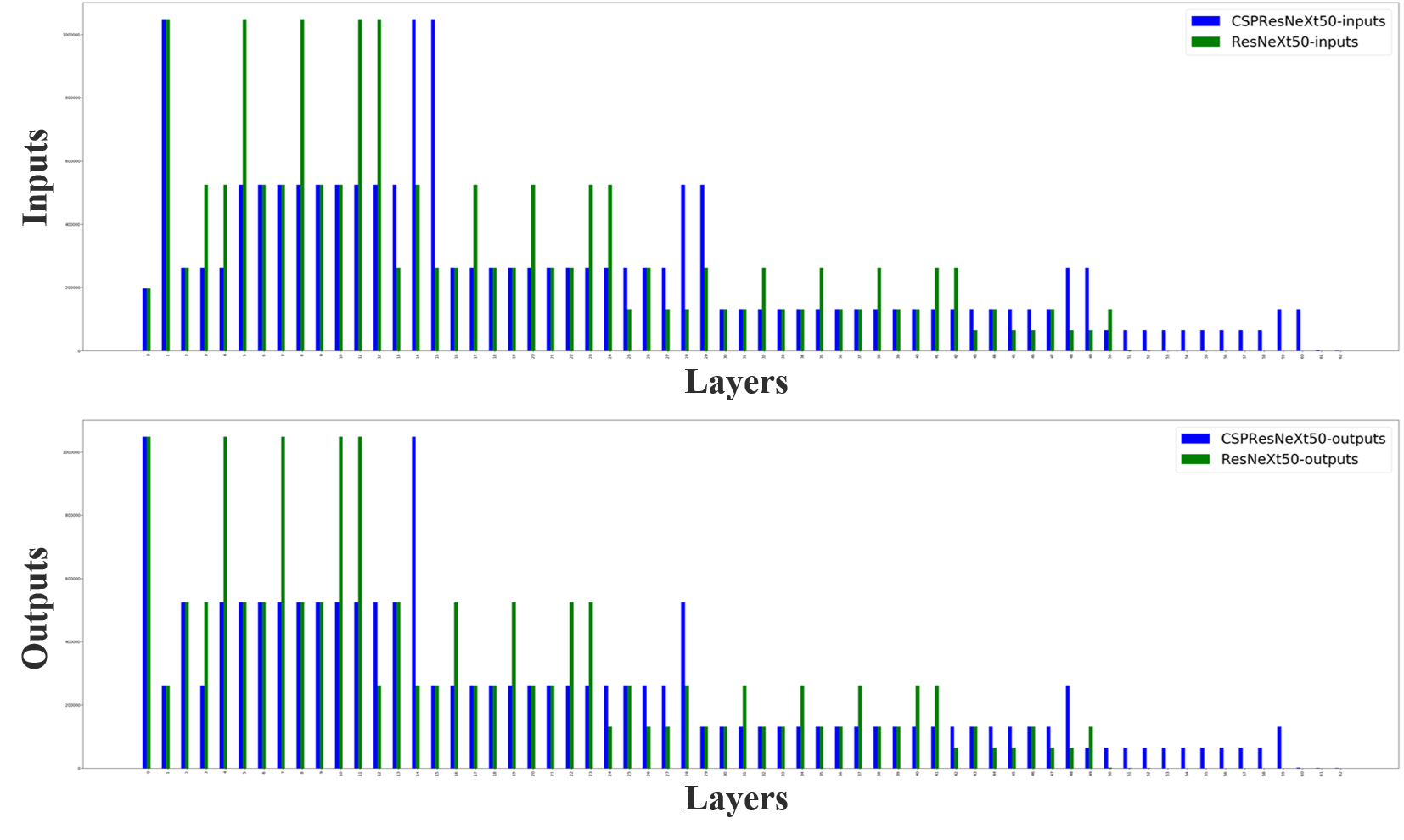}
	\end{center}
	\caption{Input size and output size of ResNeXt and proposed CSPResNeXt. }
	\label{fig:io}
\end{figure}

{\bf Memory Traffic.} Figure \ref{fig:io} shows the size of each layer of ResNeXt50 and the proposed CSPResNeXt50.  The CIO of the proposed CSPResNeXt (32.6M) is lower than that of the original ResNeXt50 (34.4M). In addition, our CSPResNeXt50 removes the bottleneck layers in the ResXBlock and maintains the same numbers of the input channel and the output channel, which is shown in Ma \textit{et al.} \cite{ma2018shufflenetv2} that this will have the lowest MAC and the most efficient computation when FLOPs are fixed. The low CIO and FLOPs enable our CSPResNeXt50 to outperform the vanilla ResNeXt50 by 22\% in terms of computations.

\vspace*{1mm}

{\bf Inference Rate.} We further evaluate whether the proposed methods are able to be deployed on real-time detectors with mobile GPU or CPU. Our experiments are based on NVIDIA Jetson TX2 and Intel Core i9-9900K, and the inference rate on CPU is evaluated with the OpenCV DNN module. We do not adopt model compression or quantization for fair comparisons. The results are shown in Table\ref{table:speed}.

\begin{table}[h]
	\centering
	\begin{threeparttable}[h]
		\footnotesize
		\caption{Inference rate on mobile GPU (mGPU) and CPU real-time object detectors (in fps).}
		\label{table:speed}
		\setlength\tabcolsep{2.0pt}
		\begin{tabular}{lccccc}
			\toprule
			Model & Size & GPU & CPU & mGPU & AP$_{50}$ \\			
			\midrule
			SNet146-Thunder \cite{qin2019thundernet} & 320 & 248 & 32 & - & 40.2 \\	
			SNet49-Thunder \cite{qin2019thundernet} & 320 & 267 & 47 & - & 33.7 \\	
			\midrule
			YOLOv3-tiny \cite{redmon2018yolov3} & 416 & 330 & 54 & 37 & 33.1 \\
			YOLOv3-tiny-PRN \cite{wang2019enriching} & 416 & \textbf{400} & 71 & 49 & 33.1 \\		
			\midrule
			\textbf{CSPPeleeNet Ref.-EFM (SAM)} & 320 & 205 & - & 41 & \textbf{44.6} \\	
			\textbf{CSPDenseNet Ref.-EFM (SAM)} & 320 & 235 & - & 49 & 42.2 \\	
			\midrule
			\textbf{CSPPeleeNet Ref.-PRN (3l)} & 320 & 267 & 52 & 38 & 40.0 \\	
			\textbf{CSPPeleeNet Ref.-PRN} & 320 & 306 & 75 & 52 & 38.5 \\	
			\textbf{CSPDenseNet Ref.-PRN} & 320 & 387 & 95 & 64 & 35.4 \\	
			\textbf{CSPDenseNetb Ref.-PRN} & 320 & \textbf{400} & \textbf{102} & \textbf{73} & 34.2 \\	
			\bottomrule
		\end{tabular}
	\end{threeparttable}
\end{table}

If we compare the inference speed executed on CPU, CSPDenseNetb Ref.-PRN receives higher AP$_{50}$ than SNet49-TunderNet, YOLOv3-tiny, and YOLOv3-tiny-PRN, and it also outperforms the above three models by 55 fps, 48 fps, and 31 fps, respectively, in terms of frame rate. On the other hand, CSPPeleeNet Ref.-PRN (3l) reaches the same accuracy level as SNet146-ThunderNet but significantly upgrades the frame rate by 20 fps on CPU.

If we compare the inference speed executed on mobile GPU, our proposed EFM will be a good model to use. Since our proposed EFM can greatly reduce the memory requirement when generating feature pyramids, it is definitely beneficial to function under the memory bandwidth restricted mobile environment. For example, CSPPeleeNet Ref.-EFM (SAM) can have a higher frame rate than YOLOv3-tiny, and its AP$_{50}$ is 11.5\% higher than YOLOv3-tiny, which is significantly upgraded. For the same CSPPeleeNet Ref. backbone, although EFM (SAM) is 62 fps slower than PRN (3l) on GTX 1080ti, it reaches 41 fps on Jetson TX2, 3 fps faster than PRN (3l), and at AP$_{50}$ 4.6\% growth.

\section{Conclusion}

We have proposed the CSPNet that enables state-of-the-art methods such as ResNet, ResNeXt, and DenseNet to be light-weighted for mobile GPUs or CPUs. One of the main contributions is that we have recognized the redundant gradient information problem that results in inefficient optimization and costly inference computations. We have proposed to utilize the cross-stage feature fusion strategy and the truncating gradient flow to enhance the variability of the learned features within different layers. In addition, we have proposed the EFM that incorporates the Maxout operation to compress the features maps generated from the feature pyramid, which largely reduces the required memory bandwidth and thus the inference is efficient enough to be compatible with edge computing devices. Experimentally, we have shown that the proposed CSPNet with the EFM significantly outperforms competitors in terms of accuracy and inference rate on mobile GPU and CPU for real-time object detection tasks.

{\small
	\bibliographystyle{ieee_fullname}  
	\bibliography{egbib-full}
}

\end{document}